\documentclass[10pt,twocolumn,a4paper]{article}

\usepackage{iccv}
\usepackage{times}
\usepackage{amsmath}
\usepackage{amssymb}
\usepackage{amstext}
\usepackage{amsthm}
\usepackage{epsfig}
\usepackage{graphicx}
\usepackage{multirow}
\usepackage{cite}
\usepackage{balance}

\usepackage[breaklinks=true,bookmarks=false]{hyperref}

\iccvfinalcopy 

\def\iccvPaperID{****} 

\begin{document}

\title{Efficient Facial Expression Analysis For Dimensional Affect Recognition\\ Using Geometric Features}

\author{Vassilios Vonikakis\thanks{This work was conducted while both authors were with the Advanced Digital Sciences Center (ADSC), University of Illinois at Urbana-Champaign, Singapore. The contribution from V. Vonikakis was made by him prior to joining AWS. Contact: winkler@comp.nus.edu.sg}\\
Amazon Web Services\\
Singapore\\
\and
Stefan Winkler\\
National University of Singapore\\
Singapore 117417\\
}

\maketitle
\ificcvfinal\thispagestyle{empty}\fi

\begin{abstract}
Despite their continued popularity, categorical approaches to affect recognition have limitations, especially in real-life situations. Dimensional models of affect offer important advantages for the recognition of subtle expressions and more fine-grained analysis.

We introduce a simple but effective facial expression analysis (FEA) system for dimensional affect, solely based on geometric features and Partial Least Squares (PLS) regression. The system jointly learns to estimate Arousal and Valence ratings from a set of facial images. The proposed approach is robust, efficient, and exhibits comparable performance to contemporary deep learning models, while requiring a fraction of the computational resources.
\end{abstract}

\section{Introduction}

Classification of prototypical high-intensity facial expressions is an extensively researched topic. Inspired initially by the seminal work of Ekman \cite{ekman1}, it has made significant strides, with increasingly powerful methods like deep learning being used \cite{bbonik_ICMI2015,emotionet}. Comprehensive reviews of this evolution can be found in \cite{review2,review1,DL_FEA}.

Despite their continued popularity, categorical emotion recognition approaches have limitations, especially in real life: people rarely exhibit high-intensity prototypical expressions in everyday situations. Most of the time, people tend to display low-key, non-prototypical expressions. As such, it is evident that other approaches are needed, such as intensity of facial action units \cite{Rudovic,AUintensity}, compound expressions \cite{compound}, or dimensional models of facial affect \cite{affectnet,AffWild}.

Russell's seminal work \cite{circumplex} has established dimensional affective models as an alternative to Ekman's categorical approach \cite{ekman1}. Neuroscientific findings \cite{psychology0} also suggest that there are separate areas in the human brain that respond to either continuous or categorical representation of emotions. This indicates that both types of  representation underlie our ability to process emotions. Interestingly though, continuous (i.e.\ dimensional) models of affect have received much less attention in the computer vision community than categorical models.

\begin{figure}[htb]
  \hspace{3mm}\includegraphics[width=0.8\columnwidth]{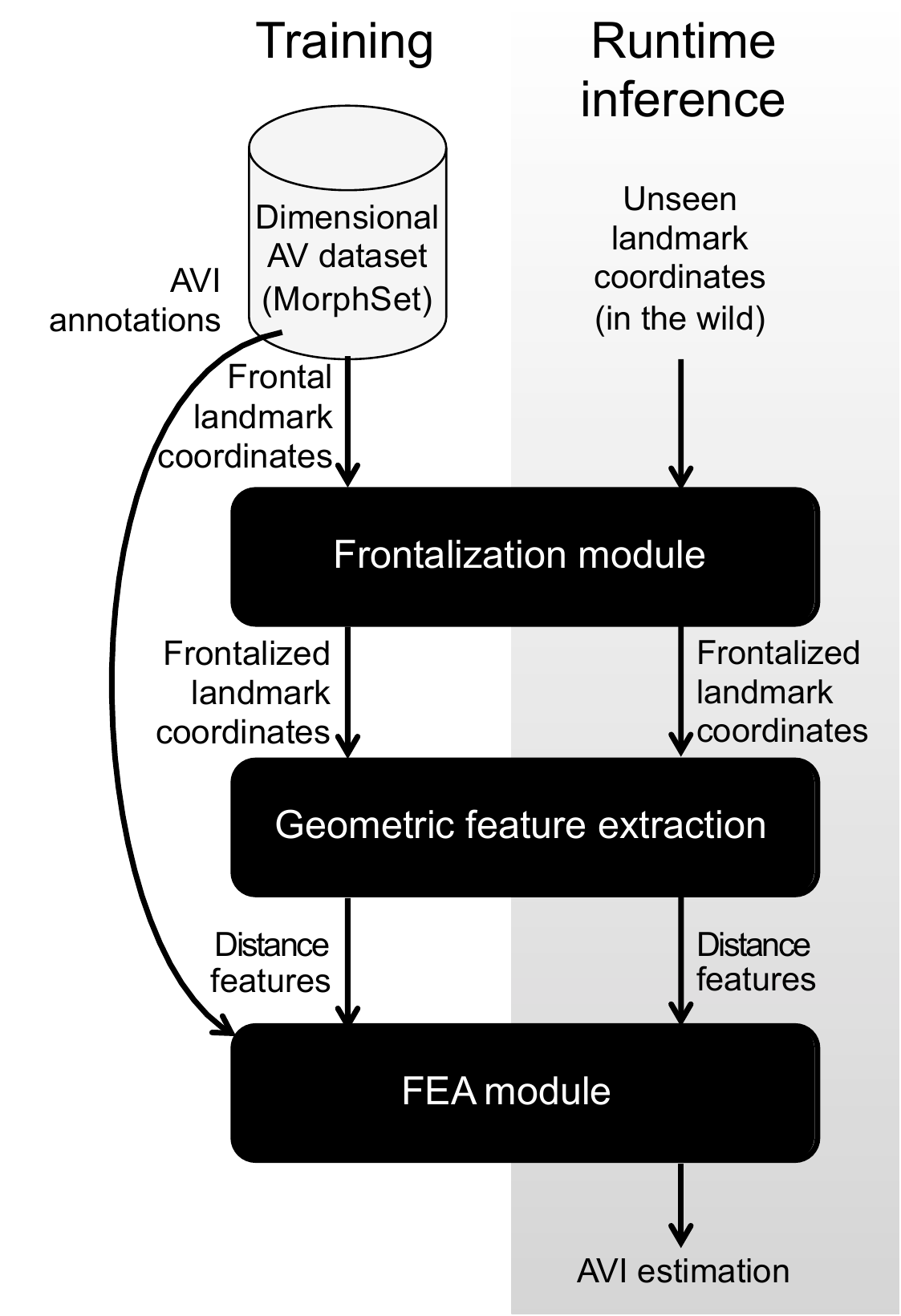}
	\caption{Block diagram of the proposed FEA system, for both training and inference phases.}
	\label{fig:overview}
\end{figure}

Building on our previous work on dataset augmentation \cite{icip2021} and face frontalization \cite{icip2020}, we introduce a simple but efficient FEA system for dimensional affect that is solely based on geometric features and Partial Least Squares (PLS) regression. The system jointly learns to estimate Arousal, Valence and Intensity from a set of facial images. The proposed technique approach comparable performance compared to contemporary deep learning models, with only a fraction of computational resources needed.  It also allows for an intuitive semantic interpretation of its features.



\section{Facial Expression Analysis}
\label{sec:FEA}

Fig.~\ref{fig:overview} depicts the block diagram of the proposed FEA system. It involves the following components:
\begin{enumerate}
    \item After face detection, Supervised Descent Method (SDM) \cite{intraface0} is used in order to estimate the 2D facial landmarks. 
    \item Invariance to head pose is essential for the system to operate on face images `in the wild'. The proposed system achieves this via landmark-based face frontalization; the approach is described in detail in \cite{icip2020}.
    \item The facial expression analysis method proper is described in Section \ref{sec:PLS} below.  It uses partial-least-squares regression from geometric features. 
    \item For training and testing, we use a dataset with dimensional Arousal, Valence, and Intensity (AVI) annotations. It is created using the `MorphSet' augmentation framework described in \cite{icip2021}.
\end{enumerate}

\subsection{Dimensional Affect}

\begin{figure}[htb]
  \centering
  \includegraphics[width=\columnwidth]{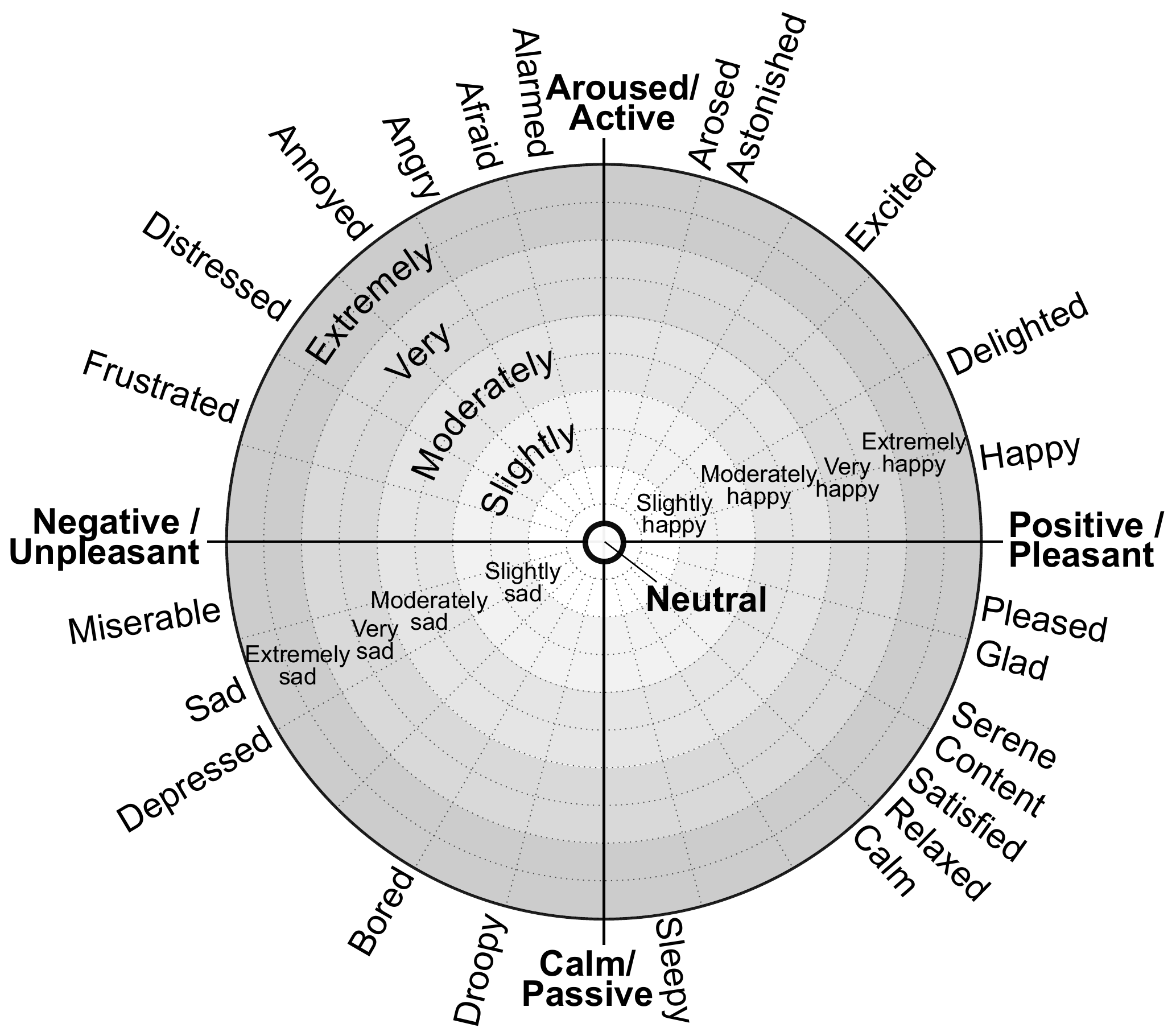}
	\caption{Illustration of the polar circumplex model of affect, adapted from \cite{circumplex}. The given intensity labels are indicative.}
    \label{fig:circumplex}
\end{figure}

We assume a 2-dimensional \emph{polar} affective space, similar to the Arousal-Valence (AV) space of the circumplex model \cite{circumplex}, with Neutral at the center. Distance from the center (i.e.\ deviation from neutral) represents the intensity of an expression. High intensity expressions (e.g.\ `extremely happy') are located at the outer perimeter of the affective space, while low-key expressions (e.g.\ `slightly happy') near the center of the space, close to Neutral. Fig.~\ref{fig:circumplex} depicts such an affective space. Arousal and Valence are in the range of $[-1,1]$.  Emotions are defined by angles in the interval $[0^{\circ},360^{\circ}]$, while intensity of expression is defined by the distance from the center (Neutral) and is in the interval $[0,1]$. We also make the following assumptions: 
\begin{itemize}
\item One-to-one correspondence of affective coordinates to facial deformations. 
\item Continuity of the facial deformation space.
\item Neighboring affective coordinates have similar facial deformations. Small changes in affective coordinates, result in small changes in the facial deformation space. 
\item Continuity of the dimensional affective space.
\end{itemize}

The above assumptions describe a mapping between emotions and a facial deformation (expression) continuum. Each set of affective coordinates corresponds to a specific facial deformation. More importantly, neighboring points in the affective continuum are also neighboring points in the facial deformation space.  


Let function $f:AV\mapsto D$ map values from the AV space to a set of facial deformations. Let also the inverse function $f^{-1}:D\mapsto AV$ map face images to the AV space. Implementing $f^{-1}$ essentially is the proposed FEA system, estimating Arousal--Valence values from face images. 

We use MorphSet \cite{icip2021}, a fast and cost-effective augmentation framework to create balanced, annotated image datasets, appropriate for training Facial Expression Analysis (FEA) systems for dimensional affect. The framework uses high-quality facial morphings to transform typical categorical datasets (usually 7 expressions per subject) into dimensional ones, with an augmentation factor of at least 20x or more.

\subsection{Frontalization}

FEA systems operating in uncontrolled conditions have to address the head pose variability issue across different identities and expressions \cite{icmi2016}. MorphSet comprises only frontal images and is thus not sufficient for training on its own. An FEA system trained only with frontal facial images would have a poor performance in `wild' conditions. 

Frontalization offer a solution to this problem by recovering the frontal view of non-frontal facial images.  
Since the proposed FEA system is based solely on geometric facial features, recovering the appearance of the frontal view of a face is not necessary. 
For this reason, we follow a \emph{landmark-based} frontalization approach, which operates only on the non-frontal facial landmark coordinates and recovers their frontal representation. This leads to a faster execution during runtime, since the computation is reduced to a simple matrix multiplication and no elaborate pixel rendering is needed. 

\begin{figure}
  \centering
  \includegraphics[width=\columnwidth]{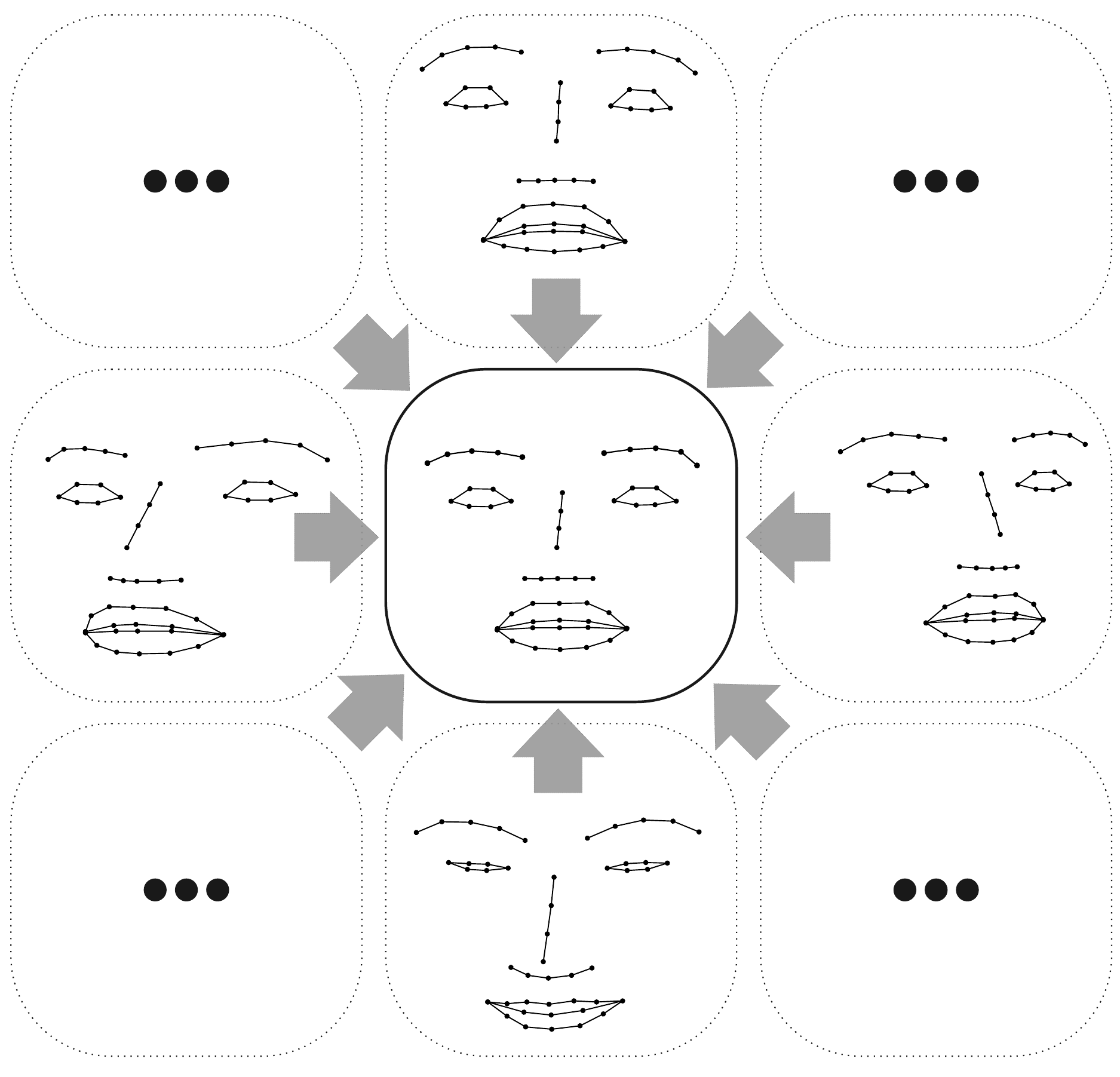}
	\caption{Illustration of the frontalization process on actual facial landmarks in a set of images from the CAS-PEAL \cite{caspeal} dataset.}
	\label{fig:frontalization}
\end{figure}

Let $\mathbf{P}^{N}_{i} \in \mathbb{R}^{N\times 2}$ be a matrix containing the $x$ and $y$ coordinates of $N$ facial landmarks of facial image $i$. These include facial points around the eyes, the nose, and the mouth. In our case, $N=49$.  The effect of the frontalization on these feature points is  illustrated in Figure~\ref{fig:frontalization}.  The  module learns a transformation that maps points from any viewpoint back to the frontal view.  The specifics of the method are described in detail in \cite{icip2020}. 

Based on our tests, the frontalization module is quite robust to yaw ranges up to $\pm 45$-$60^{\circ}$ and pitch ranges up to $\pm 15$-$30^{\circ}$.  
Faces with more extreme headposes would either yield unreliable facial landmarks or would not be detected by the face detector in the first place.

\subsection{Partial Least Squares}
\label{sec:PLS}

The standardized frontal coordinates of each facial image $i$ are used to extract facial geometric features. Similarly to \cite{distance_features}, we define these features as ``all possible combinations of Euclidean distances among the standardized coordinates of $\hat{\mathbf{P}}^{N}_{i}$''. As such, there can be $N\choose 2$ distances (in our case 1176) for each face. Although the dimensionality of these features is high, they offer a direct \emph{semantic interpretation} of facial deformations. 


The high dimensionality of the geometric feature vector can potentially be a limiting factor in the attempt to create a predictive system. Many among the 1176 distances are likely to be similar and even highly correlated, since they derive from neighboring facial landmarks. This raises the danger of multi-collinearity for typical regression-based systems. However, it is reasonable to assume that the observed data (displacement of facial landmarks / change in distances) is generated by an underlying system or process, which is driven by a smaller number of not directly observed or measured variables. 

This makes Partial Least Squares (PLS) regression very appealing for our system, since it is particularly useful for predicting a response variable from a large set of highly correlated predictors, while at the same time making use of their common structure. More specifically, PLS projects the predictors (features) to a set of orthogonal latent vectors, or \emph{components}, which have the best predictive power to approximate the response variable. In essence, it combines characteristics of both Principal Component Analysis (PCA) (maximum variance of inputs) and Ordinary Least Squares (OLS) (maximum input-output correlation), by maximizing the covariance between the response and predictor variables. As such, it performs dimensionality reduction and prediction in a single step. 

Assuming a set of predictor variables in the form of a matrix $\mathbf{X}$ (rows corresponding to observations) and a set of response variables $\mathbf{Y}$, the PLS framework decomposes them into the form:
\begin{align*}
\mathbf{X}=\mathbf{T}\mathbf{P}^\top+\mathbf{E},\\
\mathbf{Y}=\mathbf{U}\mathbf{Q}^\top+\mathbf{F},
\end{align*} 
\noindent where $\mathbf{T}$ and $\mathbf{U}$ are matrices containing the extracted latent components, $\mathbf{P}$ and $\mathbf{Q}$ represent the loadings, and $\mathbf{E}$ and $\mathbf{F}$ the residuals. The PLS algorithm finds the weight vectors $\mathbf{w}$ and $\mathbf{v}$ by optimizing the following objective function to maximize the covariance between the latent components of the predictor and the response variables:
\begin{align*}
\left[cov\left(\mathbf{t},\mathbf{u}\right)\right]^{2}=\max_{\left|\mathbf{w}\right|=\left|\mathbf{v}\right|=1}\left[cov\left(\mathbf{Xw},\mathbf{Yv}\right)\right]^2,
\end{align*} 
\noindent where $\mathbf{t}$ and $\mathbf{u}$ are the column vectors of $\mathbf{T}$ and $\mathbf{U}$, respectively, and $cov\left(\mathbf{t},\mathbf{u}\right)$ is the sample covariance. With the estimated latent components $\mathbf{T}$ and $\mathbf{U}$, the regression coefficients between $\mathbf{X}$ and $\mathbf{Y}$ is given by:
\begin{align*}
\mathbf{B}=\mathbf{X}^\top\mathbf{U}\left(\mathbf{T}^\top\mathbf{X}\mathbf{X}^\top\mathbf{U}\right)^{-1}\mathbf{T}^\top\mathbf{Y}.
\end{align*} 
As such, the predicted response can be obtained by a simple matrix multiplication $\mathbf{\hat{Y}}=\mathbf{X}\mathbf{B}$.

\section{Experiments}

The proposed FEA system is trained using Matlab's implementation of PLS called \emph{plsregress}, which is based on the SIMPLS algorithm \cite{SIMPLS}.

During training of the FEA module, all training examples are passed though the frontalization module, even though they are all frontal. This is done because the frontalization step inevitably introduces small changes even to frontal images. These changes can introduce variations that may compromise the performance of the FEA system. By passing all frontal training images through the frontalization module, the system is able to adapt to these small changes and learn to predict the correct affective values of the whole pipeline.

\subsection{Latent Components}

The number of latent components plays an important role in the success of the model and may act as a kind of regularization. Deciding on the number of components is very important. A large number of components will do a good job in fitting the current observed data, but may result in overfitting and thus poor generalization. For this reason we studied the impact of different numbers of latent components on the overall predictive performance. 

As such, we split the dataset into 2 disjoint parts; training and validation. For the validation, we select a total of 20 subjects (10 males and 10 females) from the dataset.  In total, the validation set comprises approximately 8\% of the total size of the overall dataset.

\begin{figure}[htb]
  \centering
  \includegraphics[width=0.49\textwidth]{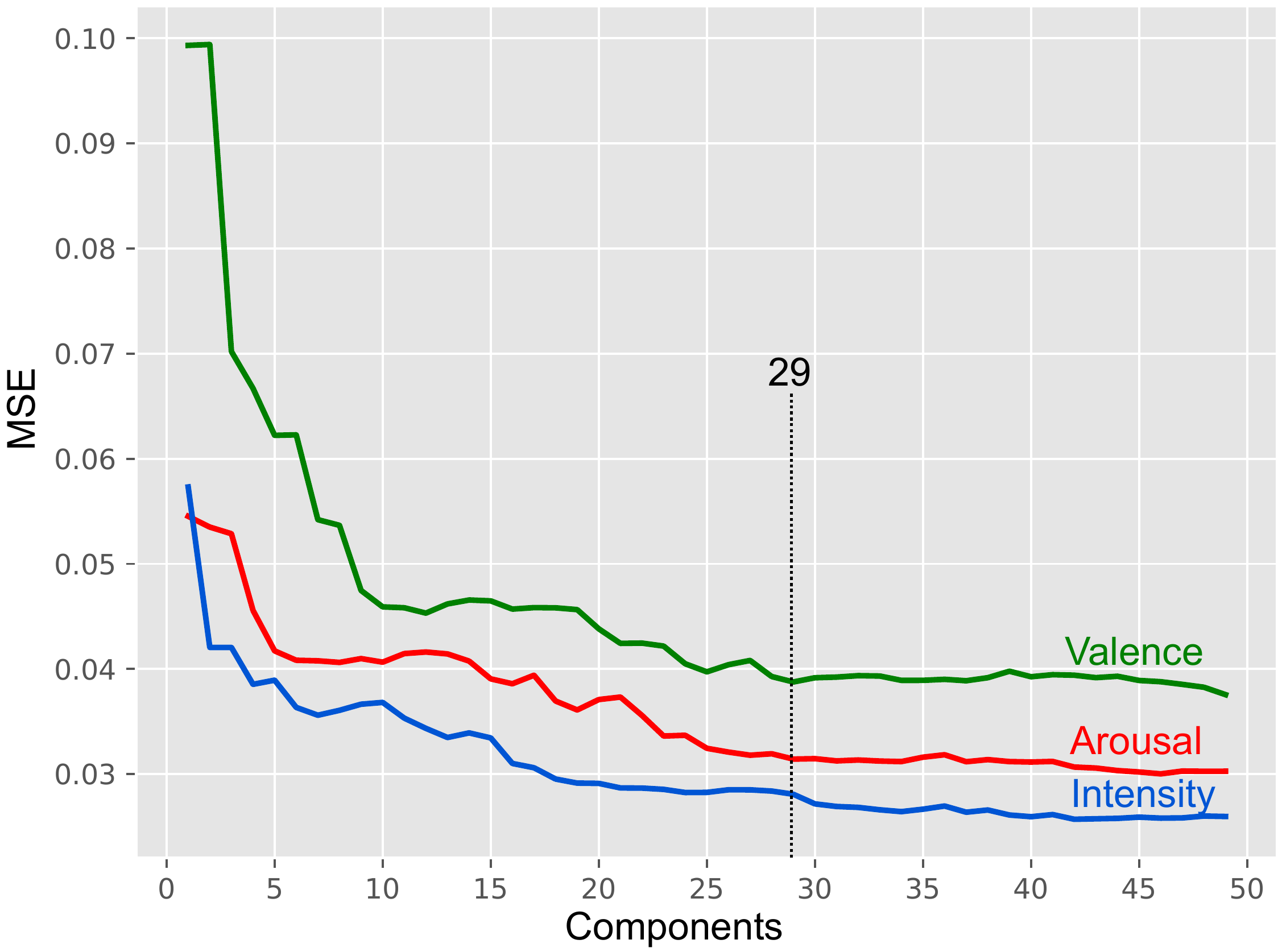}
	\caption{MSE of the FEA system for valence, arousal, and intensity estimation on the validation dataset. The vertical line shows the selected number of components (i.e.\ 29).}
	\label{fig:components}
\end{figure}

The objective here is to select the \emph{smallest} number of latent components (more regularized model) that achieves good accuracy. Fig.~\ref{fig:components} depicts the Mean Squared Error (MSE) of the trained FEA system for the validation dataset, for different numbers of latent components.  Approximately every 10 additional components, there is a drop in the validation MSE. From about 30 components and above, the curves remain relatively flat, and MSE changes are minimal. 
Based on this, we select 29 latent components for the final FEA system, which achieves the best balance between regularization and predictive performance. 

Interestingly, the selected number of latent components is very close to the number of the main Action Units (i.e.\ 28) that describe all \emph{basic} facial actions in the Facial Action Coding System (FACS) \cite{FACS}. This indicates that the PLS optimization captures some internal facial structure, similar in complexity to the main Action Units. 

\subsection{Regression Alternatives}
Apart from PLS, we also explore other regression methods, including Ridge regression, Principal Components Regression (PCR) (PCA combined with least squares), Random Forest regression (RF), and multitask Multi-Layer Perceptron (MLP). Table \ref{tab:regressions} lists the performance of these methods. 

MLP with 3 hidden layers exhibits the best MSE performance for Arousal, Ridge regression the best for Valence, and PLS with 100 components the best for Intensity. Overall, the performance is comparable to a baseline deep convolutional neural network (ResNet-18) \cite{icip2021}.

\begin{table}[htb]
\centering
\caption{MSE of different regression methods discussed in the text.  Model parameters are shown in parentheses. Lowest values in each column are highlighted in bold.}
\begin{tabular}{lccc}
\hline
Method                  & Arousal                     & Valence                              & Intensity                            \\ \hline
MLP (100,50)       & 0.0350                      & 0.0526                               & 0.0319                               \\ 
MLP (50,50,50)     & \textbf{0.0276}             & 0.0412                               & 0.0275                               \\ 
RF (10)           & 0.0351                      & 0.0438                               & 0.0329                               \\ 
RF (50)           & 0.0304                      & 0.0410                               & 0.0301                               \\ 
Ridge ($\alpha=0.1$)     & 0.0301                      & {\textbf{0.0367}}                               & 0.0237                               \\ 
Ridge ($\alpha=0.5$)     & {0.0302} & {0.0371} & {0.0253}          \\
PCR (29)          & {0.0381} & {0.0452}          & {0.0338}          \\ 
PCR (100)         & {0.0305} & {0.0382}          & {0.0267}          \\ 
\textit{PLS (29)}          & \textit{0.0314} & \textit{0.0388}          & \textit{0.0281}          \\ 
PLS (100)         & {0.0321} & {0.0378}          & {\textbf{0.0235}} \\ \hline
\end{tabular}
\label{tab:regressions}
\end{table}

\subsection{Qualitative Testing in the Wild}

Apart from the quantitative testing shown above, all regression methods were also tested in real-time with a live feed from a web camera on unseen faces and in varying illumination conditions. Interestingly, the methods with the most robust performance during this \emph{qualitative} testing are not necessarily the ones with the smallest MSE error in Table \ref{tab:regressions}. Specifically, we observed that dimensionality reduction methods such as PLS and PCR, are much more stable when tested with totally different data distributions (like a web camera feed), compared to the others. This essentially substantiates a better generalization behavior. Between the two, PLS outperforms PCR, achieving a more regularized model with fewer components. 

Based on the above findings, PLS with 29 components is selected for the final FEA system. Although PLS(29) does not achieve the lowest error individually, in any of the 3 objectives, it is  consistently among the better methods,  making it a very good choice overall. A possible explanation for this behavior is that the latent components learned by PLS capture some internal structure of the facial deformations, resulting in more regularized models with better generalization characteristics.

\subsection{Feature Interpretation}

\begin{figure}[htb]
  \centering
  \includegraphics[width=\columnwidth]{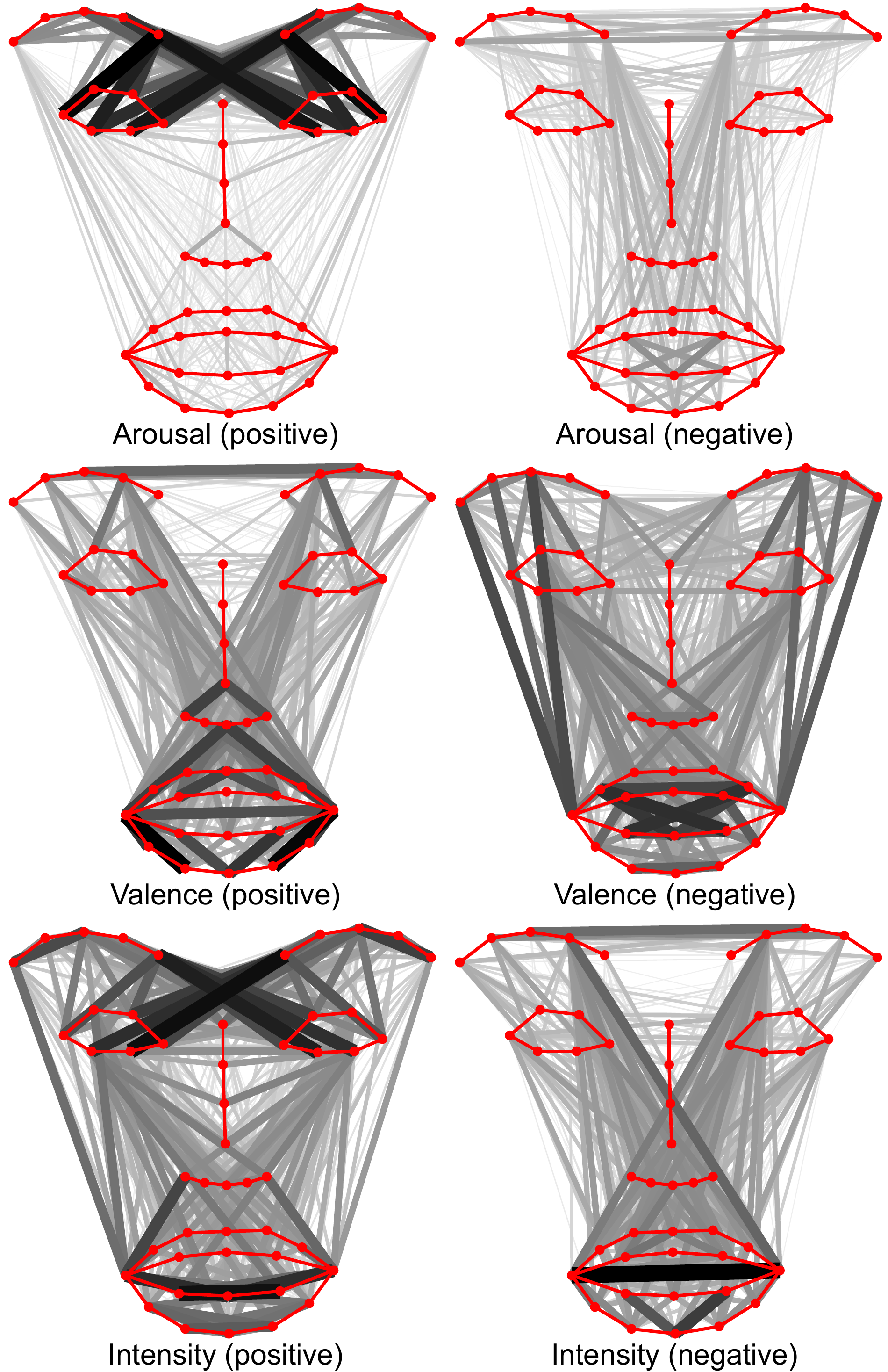}
	\caption{Top 500 positive and negative weights (distances) for Arousal, Valence and Intensity, learnt using the PLS approach. Thicker and darker lines indicate larger weights.}
	\label{fig:AVIweights}
\end{figure}

The PLS weights can give some important insights on what the model learns. At the same time, the fact that only distances between facial landmarks are used as features, allows an intuitive interpretation on what the model associates with its outputs. Fig.~\ref{fig:AVIweights} depicts the top 500 positive and negative distances (out of the total 1176), which contribute positively and negatively to the final predicted values. Positive distances essentially indicate that, when increased (through facial deformations), they will drive the estimated affective value up. Conversely, when negative distances are increased, they drive the estimated affective value down. 

Distances that encode the position of the eyebrows in relation to the eyes, seem to be associated more with high Arousal. The inner parts of the eyebrow seem to be more important than the outer parts. One would expect that eye opening should also contribute to the increase of Arousal. Note that the dataset (MorphSet) does not include training images with high negative Arousal (e.g.\ sleepy, drowsy etc.) because of the underlying basic emotions. 

Increases in Valence are linked to lengthening of distances that encode the eccentricity of the mouth, especially in relation to the mouth corners. Additionally, increasing the distance between the middle of the eyebrows, as well as, slightly lifting them (relatively to the eyes), also contributes to higher Valence. Interestingly, this coincides with previous reports about discriminating genuine from fake smiles, showing that the major difference lies in muscular activity around eyes \cite{genuine_smiles}. On the other hand, distances that lead to a squarish opening of the mouth, along with increasing the outer corner--to--corner distances between mouth--eyebrow pairs (either by lowering the mouth corners, or raising the outer eyebrow corners), are associated with negative Valence. 

As expected, Intensity seems to be a combination of the characteristics of Arousal and Valence. Again, the inner part of the eyebrows, in relation to the eyes, seems to be important, along with the eccentricity of the mouth. We also see that distances associated with positive and negative Valence/Arousal are combined together. For example, the positive Arousal eyebrow distances are combined with the negative Valence outer corner--to--corner mouth--eyebrow distances, encoding increased Intensity. This is expected, since Intensity of expression encodes the \emph{magnitude} of deformation, irrespective of whether this derives from positive or negative Arousal or Valence.

\section{Conclusion}
\label{sec:conclusions}
We presented a system for facial expression analysis (FEA) to estimate dimensional emotion labels in the arousal/valence (AV) space. The proposed approach uses 49 facial landmarks for both frontalization and AVI estimation. The system is trained on MorphSet \cite{icip2021} and tested on live camera feeds in the wild.  Our approach is computationally efficient and robust to varying environmental conditions. Additionally, the learnt geometrical features are easy to interpret and appear to be in accordance with  psychophysical observations about affect.

\balance
\bibliographystyle{IEEEtran}
\bibliography{short,refs}

\end{document}